\documentclass{article}
\usepackage{graphicx} 
\usepackage[T1]{fontenc}
\usepackage[latin1]{inputenc}
\usepackage{float} 
\usepackage{graphicx}
\usepackage[margin=1in]{geometry}
\usepackage{setspace}
\usepackage{amsmath,amssymb,amsfonts}
\usepackage{acronym}
\usepackage{units}
\usepackage{hyperref}
\usepackage{fancyhdr}
\usepackage{wrapfig}
\usepackage{caption}
\usepackage{subfig}
\usepackage{multirow}
\usepackage{sidecap}
\usepackage{xcolor}
\usepackage{lineno}
\usepackage{xr}
\usepackage{diagbox}
\usepackage{authblk}
\usepackage{tikz}
\usepackage{makecell}

\definecolor{red}{rgb}{1.0,0.0,0.0}

\newcommand{\refeq}[1]{Eq.~(\ref{#1})}
\newcommand{\reftab}[1]{Tab.~\ref{#1}}
\newcommand{\reffig}[1]{Fig.~\ref{#1}}

\newcommand*{\corresponding}[1]{%
    \normalsize\href{mailto:#1}{#1}\par
    }

\date{}

\begin{document}

\title{Emulating Brain-like Rapid Learning in Neuromorphic Edge Computing}

\author{Kenneth Stewart$^{1,4}$, Michael Neumeier$^2$, Sumit Bam Shrestha$^3$, Garrick Orchard$^3$, Emre Neftci$^4$}

\affil{$^1$ Department of Computer Science, UC Irvine,
Irvine California, USA}
\affil{$^2$ Department of Neuromorphic Computing, Fortiss-Research Institute of the Free State of Bavaria, Munich, Germany}
\affil{$^3$ Intel Labs, Intel Corporation,
Santa Clara, California}
\affil{$^4$ Peter Gr\"unberg Institute -- Neuromorphic Software Ecosystems\\ Forschungszentrum J\"ulich, Germany\\ Faculty of Electrical Engineering and Information Technology, RWTH Aachen, Germany\\ \corresponding{kennetms@uci.edu, e.neftci@fz-juelich.de}}

\acrodef{IR}[IR]{Intrinsic Rewards and Motivation}
\acrodef{PPO}[PPO]{Proximal Policy Optimization}
\acrodef{RL}[RL]{Reinforcement Learning}
\acrodef{AC}[AC]{Arrenhius \& Current}
\acrodef{AD}[AD]{Automatic Differentiation}
\acrodef{AER}[AER]{Address Event Representation}
\acrodef{AEX}[AEX]{AER EXtension board}
\acrodef{AMDA}[AMDA]{``AER Motherboard with D/A converters''}
\acrodef{ANN}[ANN]{Artificial Neural Network}
\acrodef{API}[API]{Application Programming Interface}
\acrodef{BP}[BP]{Back-Propagation}
\acrodef{BPTT}[BPTT]{Back-Propagation-Through-Time}
\acrodef{BM}[BM]{Boltzmann Machine}
\acrodef{CAVIAR}[CAVIAR]{Convolution AER Vision Architecture for Real-Time}
\acrodef{CCN}[CCN]{Cooperative and Competitive Network}
\acrodef{CD}[CD]{Contrastive Divergence}
\acrodef{CG}[CG]{Computational Graph}
\acrodef{CMOS}[CMOS]{Complementary Metal--Oxide--Semiconductor}
\acrodef{CNN}[CNN]{Convolutional Neural Network}
\acrodef{COTS}[COTS]{Commercial Off-The-Shelf}
\acrodef{CPU}[CPU]{Central Processing Unit}
\acrodef{CV}[CV]{Coefficient of Variation}
\acrodef{CTC}[CTC]{connectionist temporal classification}
\acrodef{DAC}[DAC]{Digital--to--Analog}
\acrodef{DBN}[DBN]{Deep Belief Network}
\acrodef{DCLL}[DECOLLE]{Deep Continuous Local Learning}
\acrodef{DFA}[DFA]{Deterministic Finite Automaton}
\acrodef{DFA}[DFA]{Deterministic Finite Automaton}
\acrodef{divmod3}[DIVMOD3]{divisibility of a number by 3}
\acrodef{DPE}[DPE]{Dynamic Parameter Estimation}
\acrodef{DPI}[DPI]{Differential-Pair Integrator}
\acrodef{DSP}[DSP]{Digital Signal Processor}
\acrodef{DVS}[DVS]{Dynamic Vision Sensor}
\acrodef{EDVAC}[EDVAC]{Electronic Discrete Variable Automatic Computer}
\acrodef{EIF}[EI\&F]{Exponential Integrate \& Fire}
\acrodef{EIN}[EIN]{Excitatory--Inhibitory Network}
\acrodef{EPSC}[EPSC]{Excitatory Post-Synaptic Current}
\acrodef{EPSP}[EPSP]{Excitatory Post--Synaptic Potential}
\acrodef{eRBP}[eRBP]{Event-Driven Random Back-Propagation}
\acrodef{FPGA}[FPGA]{Field Programmable Gate Array}
\acrodef{FSM}[FSM]{Finite State Machine}
\acrodef{GPU}[GPU]{Graphical Processing Unit}
\acrodef{HAL}[HAL]{Hardware Abstraction Layer}
\acrodef{HH}[H\&H]{Hodgkin \& Huxley}
\acrodef{HMM}[HMM]{Hidden Markov Model}
\acrodef{HW}[HW]{Hardware}
\acrodef{hWTA}[hWTA]{Hard Winner--Take--All}
\acrodef{IF2DWTA}[IF2DWTA]{Integrate \& Fire 2--Dimensional WTA}
\acrodef{IF}[I\&F]{Integrate \& Fire}
\acrodef{IFSLWTA}[IFSLWTA]{Integrate \& Fire Stop Learning WTA}
\acrodef{INCF}[INCF]{International Neuroinformatics Coordinating Facility}
\acrodef{INRC}[INRC]{Intel Neuromorphic Research Community}
\acrodef{INI}[INI]{Institute of Neuroinformatics}
\acrodef{IO}[IO]{Input-Output}
\acrodef{IoT}[IoT]{internet of things}
\acrodef{IPSC}[IPSC]{Inhibitory Post-Synaptic Current}
\acrodef{ISI}[ISI]{Inter--Spike Interval}
\acrodef{JFLAP}[JFLAP]{Java - Formal Languages and Automata Package}
\acrodef{LIF}[LIF]{Linear Integrate and Fire}
\acrodef{LSM}[LSM]{Liquid State Machine}
\acrodef{LTD}[LTD]{Long-Term Depression}
\acrodef{LTI}[LTI]{Linear Time-Invariant}
\acrodef{LTP}[LTP]{Long-Term Potentiation}
\acrodef{LTU}[LTU]{Linear Threshold Unit}
\acrodef{LSTM}[LSTM]{Long Short-Term Memory}
\acrodef{MAML}[MAML]{Model Agnostic Meta Learning}
\acrodef{MCMC}{Markov Chain Monte Carlo}
\acrodef{MSE}{Mean-Squared Error}
\acrodef{NHML}[NHML]{Neuromorphic Hardware Mark-up Language}
\acrodef{NMDA}[NMDA]{NMDA}
\acrodef{NME}[NE]{Neuromorphic Engineering}
\acrodef{PCB}[PCB]{Printed Circuit Board}
\acrodef{PRC}[PRC]{Phase Response Curve}
\acrodef{PSC}[PSC]{Post-Synaptic Current}
\acrodef{PSP}[PSP]{Post--Synaptic Potential}
\acrodef{RI}[KL]{Kullback-Leibler}
\acrodef{RRAM}[RRAM]{Resistive Random-Access Memory}
\acrodef{RBM}[RBM]{Restricted Boltzmann Machine}
\acrodef{RTRL}[RTRL]{Real-Time Recurrent Learning}
\acrodef{ROC}[ROC]{Receiver Operator Characteristic}
\acrodef{RSA}[RSA]{Representational Similarity Analysis}
\acrodef{RDA}[RDA]{Representational Dissimilarity Analysis}
\acrodef{RDM}[RDA]{Representational Dissimilarity Matrix}
\acrodef{RNN}[RNN]{Recurrent Neural Network}
\acrodef{SAC}[SAC]{Selective Attention Chip}
\acrodef{SCD}[SCD]{Spike-Based Contrastive Divergence}
\acrodef{SCX}[SCX]{Silicon CorteX}
\acrodef{SG}[SG]{Surrogate Gradient}
\acrodef{SGD}[SGD]{Stochastic Gradient Descent}
\acrodef{SRM}[SRM]{Spike Response Model}
\acrodef{SNN}[SNN]{Spiking Neural Network}
\acrodef{STDP}[STDP]{Spike Time Dependent Plasticity}
\acrodef{SW}[SW]{Software}
\acrodef{sWTA}[SWTA]{Soft Winner--Take--All}
\acrodef{TPU}[TPU]{Tensorflow Processing Unit}
\acrodef{VHDL}[VHDL]{VHSIC Hardware Description Language}
\acrodef{VLSI}[VLSI]{Very  Large  Scale  Integration}
\acrodef{WTA}[WTA]{Winner--Take--All}
\acrodef{XML}[XML]{eXtensible Mark-up Language}

\acrodef{CUBA}[CUBA]{Current Based}
\acrodef{SOEL}[SOEL]{Surrogate-gradient Online Error-triggered Learning}

\maketitle 


\begin{abstract}
Achieving personalized intelligence at the edge with real-time learning capabilities holds enormous promise in enhancing our daily experiences and helping decision making, planning, and sensing. 
However, efficient and reliable edge learning remains difficult with current technology due to the lack of personalized data, insufficient hardware capabilities, and inherent challenges posed by online learning.

Over time and across multiple developmental stages, the brain has evolved to efficiently incorporate new knowledge by gradually building on previous knowledge. 
In this work, we emulate the multiple stages of learning with digital neuromorphic technology that simulates the neural and synaptic processes of the brain using two stages of learning.
First, a meta-training stage trains the hyperparameters of synaptic plasticity for one-shot learning using a differentiable simulation of the neuromorphic hardware.
This meta-training process refines a hardware local three-factor synaptic plasticity rule and its associated hyperparameters to align with the trained task domain.
In a subsequent deployment stage, these optimized hyperparameters enable fast, data-efficient, and accurate learning of new classes.
We demonstrate our approach using event-driven vision sensor data and the Intel Loihi neuromorphic processor with its plasticity dynamics, achieving real-time one-shot learning of new classes that is vastly improved over transfer learning.
Our methodology can be deployed with arbitrary plasticity models and can be applied to situations demanding quick learning and adaptation at the edge, such as navigating unfamiliar environments or learning unexpected categories of data through user engagement.

\end{abstract}

\section{Introduction}

The brain's learning strategies differ remarkably with deep neural networks in that learning is local and largely incompatible with gradient backpropagation, and it does not learn tasks from scratch. 
Throughout evolution, infancy, and adulthood, brains acquire inductive biases to survive and thrive.
Even before birth, brains are specialized to solve tasks that are likely to be encountered in their lifetimes and environment \cite{Sterling_Laughlin15_prinneur}. 
Furthermore, they are equipped with restricted but highly optimized mechanisms to adapt to new variations and situations.

In this work, we take inspiration from these observations to lay out an algorithmic framework that equips neuromorphic hardware with optimized mechanisms for rapid and efficient learning at the edge.
Neuromorphic electronic systems aim to mimic the structure and dynamics of the brain to facilitate efficient, versatile, and rapid information processing \cite{Mead90_neurelec,Indiveri_etal11_neursili,Friedmann_etal17_demohybr,Davies_etal18_loihneur,Benjamin_etal14_neurmixe}.
Neuromorphic hardware generally implements \acp{SNN}, which exploit temporal sparsity by means of asynchronous, event-based communication \cite{Lazzaro_etal93_siliaudi,Chicca_etal13_neurelec}.
Neuromorphic \emph{learning} capabilities further enable rapid and energy-efficient learning at the edge via local synaptic plasticity rules \cite{cartiglia2022onlinemix,Stewart_etal20_onchfews,Chicca_etal13_neurelec,Arthur_Boahen06_learsili}.
Prior work with neuromorphic systems demonstrated learning with three-factor synaptic plasticity from streaming data to incorporate new information, enabling privacy-preserving and low-latency adaptation to new data 
\cite{stewart2022meta,cartiglia2022onlinemix,Chicca_etal13_neurelec} at accuracies on par with deep neural networks.
However, prior approaches did not deliver viable online learning technology, due to long training times, the temporally correlated and real-time nature of the data, or ineffective learning rules compared to their mathematically derived counterparts \cite{Zenke_Neftci21_brailear}.

\begin{figure}
    \centering
    \includegraphics[width=\textwidth]{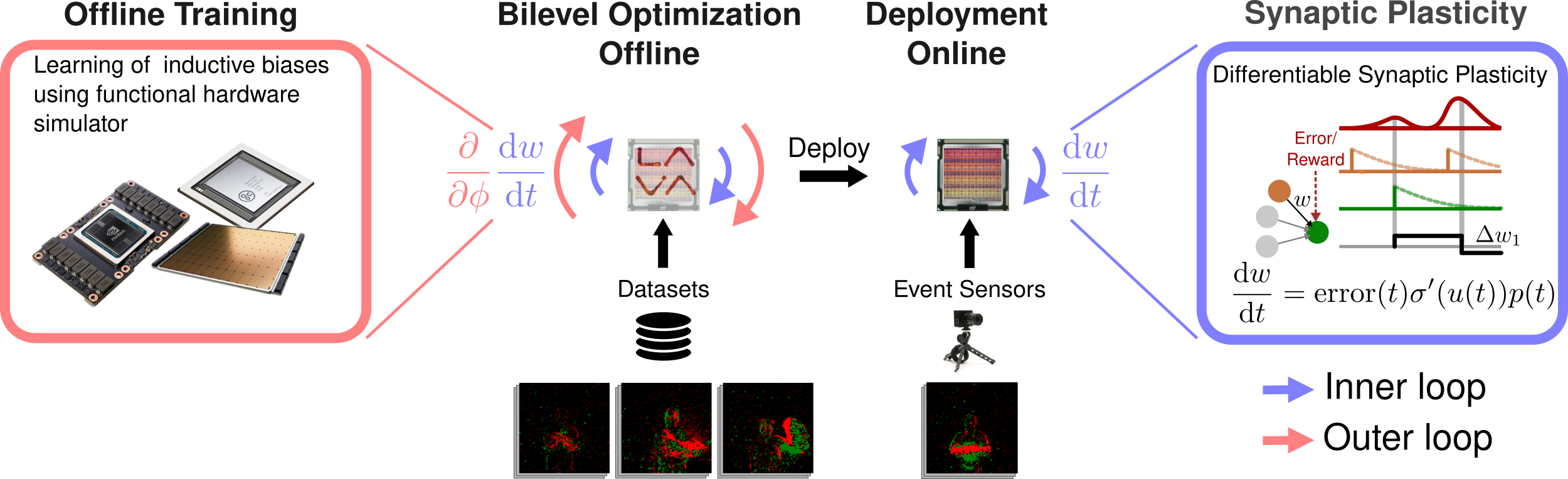}
    \caption{Our approach consists of two stages of learning whereby an offline stage uses off-the-self deep learning accelerators and GPUs to optimize the hyperparameters $\phi$ of a simulation of the learning hardware, and a deployment stage where the online weights $w$ are tuned to solve a new task. This bi-level learning setup can be viewed as a method to inject the inductive biases in a given task domain (\emph{e.g.} gestures recorded with event-based vision sensors) necessary to facilitate the learning of new tasks (e.g. new gestures and subjects).}
    \label{fig:snn_loihi}
\end{figure}

Our approach solves these problems by distributing the learning process into two stages (\reffig{fig:snn_loihi}).
The first stage uses a cycle-accurate differentiable simulator of a target digital neuromorphic learning hardware to pre-train the plasticity and learning dynamics. 
Its differentiable nature allows for the computation of model parameter and hyperparameter gradients across the learning trajectory.
This procedure is a type of \emph{bi-level optimization} \cite{Blondel_etal21_effimodu}, where an outer loop optimizes over tasks sampled from a selected task, and where the inner loop dynamics captures the learning process on the sampled tasks.
This bilevel optimization is achieved here using a variation of the \ac{MAML} algorithm \cite{finn2018metalearning,Miconi_etal20_backtrai}, where gradients are computed across the plasticity trajectory.
The second stage is the deployment: The model trained in the first stage (called henceforth the pre-trained model) is deployed on the target hardware to learn from streamed event data.

The first stage is accomplished offline with high-throughput, flexible computers and a large dataset, while the second stage can take place online with power-efficient and real-time hardware using streaming inputs. 
This distributed approach shifts the most energy and data intensive stages of learning to data centers, away from the edge where power and chip area are limited.
Bi-level optimization has a high energy and time cost since it optimizes the \emph{learning} trajectory on a large number of tasks. 
However, in its intended applications, the cost can be amortized by using the same pre-trained model on a large number of learning devices at the edge.
Transferring pre-trained learning models to hardware has been previously achieved in \acp{SNN} using surrogate gradient methods \cite{Neftci_etal19_surrgrad} for digital \cite{Stewart_etal20_onlifews} and mixed-signal \cite{Cramer_etal22_surrgrad} neuromorphic hardware.
Prior work has also demonstrated such bi-level co-optimization with software-simulated \acp{SNN} and surrogate gradients on few-shot learning problems \cite{stewart2022meta}. 
However, the ability to transfer differentiable synaptic plasticity models for \emph{online} learning with neuromorphic hardware has not yet been achieved.

Here, we take an important step beyond prior work by meta-training for digital neuromorphic hardware deployment, and demonstrating this in the case of the Intel Loihi Research Chip (``Loihi'') and its on-chip synaptic plasticity.
Bi-level optimization is used to train network hyperparameters with the specific plasticity dynamics of the neuromorphic hardware and target task domain.
Specifically, we use \ac{SOEL} \cite{Stewart_etal20_onlifews}, a local three-factor synaptic plasticity learning rule previously demonstrated for transfer learning in neuromorphic hardware \cite{Stewart_etal20_onchfews,Payvand_etal20_oncherro}.
The three-factor synaptic plasticity rules used in this work are function of local information, namely from pre- and post-synaptic connections, and a global error used to optimize the synapses towards and objective. 
Three factor rules are consistent with the dynamics of biological synapses and constitute a normative theory of learning in the brain \cite{Gerstner_etal18_eligtrac}.
Using local information, less information is communicated on-chip \cite{Chicca_etal13_neurelec,Davies_etal18_loihneur} enabling highly efficient adaptation to streamed sensory data.
Furthermore, this leads to low latency since the update locking is drastically reduced \cite{Jaderberg_etal16_deconeur}.

In the following sections we describe the underlying plasticity dynamics and its implementation in digital neuromorphic hardware. Then, we demonstrate its use within the bi-level optimization with meta-learning using Intel's neuromorphic hardware as a case study.
\section{Methods}
We use bi-level optimization to train the parameters of SNNs equipped with plasticity dynamics on event-based vision tasks domains.
Learning and synaptic plasticity in neuromorphic hardware is challenging because gradient-based learning stands in contradiction with the online and local nature of synaptic plasticity, namely, by violating the required i.i.d. assumptions for neural network convergence.
Therefore, rather than learning solely on neuromorphic hardware, a typical approach is to pre-train a network offline in general-purpose hardware, deploy it on-chip and perform learning and adaptation to streaming data online.
To facilitate this approach, Intel developed the Lava-DL\footnote{Lava-DL is a deep SNN training and inference framework available publicly at \url{https://github.com/lava-nc/lava-dl}} SNN training library, which includes a differentiable simulation of the Intel Loihi neurons with the same constraints (\emph{e.g.} parameter quantization, resource limits) as the hardware.
To demonstrate the meta-training of a synaptic plasticity rule for implementation on Intel's Loihi neuromorphic hardware, we demonstrate bi-level learning on an error-triggered synaptic plasticity previously demonstrated on the Loihi neuromorphic hardware \cite{Payvand_etal20_errothre,Stewart_etal20_onlifews}.
Specifically, we demonstrate a differentiable and quantized implementation of the Surrogate-gradient Online Error-triggered  Learning (SOEL) rule first meta-trained in Lava-DL and then mapped to the Loihi hardware for further adaptation with online learning.

\subsection{\acl{SOEL} in Digital Neuromorphic Hardware}
In this work, we focus on digital neuromorphic processors equipped with plasticity dynamics \cite{Frenkel_Indiveri22_reck28nm,Detorakis_etal18_neursyna,Davies_etal18_loihneur}, namely the the Intel Loihi which equipped with neuromorphic cores capable of on-chip synaptic plasticity \cite{Davies_etal18_loihneur,orchard2021efficient}.
 We utilize a plasticity rule adapted from a three-factor rule called \acf{SOEL} \cite{Stewart_etal20_onlifews,Stewart_etal20_onchfews}, which has been successively demonstrated in other few-shot learning scenarios \cite{Stewart_etal20_onlifews,Stewart_Neftci22_metaspik}. 
\ac{SOEL} features three key properties that make it an excellent candidate for efficient synaptic plasticity rule on digital neuromorphic hardware: (i) Surrogate gradients, (ii) Multifactors, and (iii) Error-triggering, each of which is explained below.
(i) The surrogate gradient approach solves the non-differentiability problem of the spiking neuron \cite{Neftci_etal19_surrgrad} by using a differentiable surrogate of the network for gradient computation. 
For the purpose of computing the gradient only, this surrogate network typically replaces the hard threshold firing of the \ac{LIF} with a Sigmoid-like function.
Due to its simplicity, excellent practical performance, and the ability to leverage existing machine learning hardware and software, the surrogate gradient method has become the \textit{de facto} method for computing the gradients of \acp{SNN} \cite{stewart2022meta,Cramer_etal20_traispik,Bellec_etal19_biolinsp,Bohnstingl_etal20_onlispat,Zenke_Ganguli18_supesupe,Neftci_etal17_evenranda}. 

(ii) The multi-factor aspect enables an efficient implementation in hardware. 
The analytical form of the synaptic weight gradients in a single \ac{LIF} network reveals three factors, a global error signal, a local post-synaptic factor, and a local pre-synaptic factor \cite{Zenke_Ganguli18_supesupe,Neftci_etal19_surrgrad}:
\begin{equation}\label{eq:loss}
\Delta w_{ij}(t) = 
\frac{\partial \mathcal{L} (s(t))}{\partial s_{i}(t)} 
\frac{\partial s_i(t)          }{\partial u_{i}(t)} 
\frac{\partial u_i(t)          }{\partial w_{ij}} =: \text{error}(t) \sigma'(u_i(t)) p_j(t),
\end{equation}
where $\mathcal{L}$ is a differentiable loss function, $u_i(t)$ is the membrane potential of the neuron $i$ at time t, $w_{ij}$ is the weight of the synapse connection between neurons $j$ and $i$, and $p_j(t)$ is the trace of the pre-synaptic input from neuron $j$. 
The term $\sigma'$ is the gradient of the activation function of the neuron. 
Because spiking neurons are not differentiable due to their sharp firing threshold, the naive analytical $\sigma'$ is zero everywhere except at the origin, where it is infinity. 
The surrogate gradient approach mitigates this problem by replacing this term with a differentiable one, such as the derivative of the sigmoid \cite{Zenke_Ganguli18_supesupe} or a box-car function \cite{Payvand_etal20_errothre}.
The first factor, $\text{error}(t)$, acts as an error-dependent external modulation of the weight update. 
The factor $p_j$ is related to the pre-synaptic trace used in the classical \ac{STDP} rules, making its implementation compatible with most existing neuromorphic learning hardware.
Furthermore, only one trace per neuron is required (there are no terms with indices $i$ and $j$ simultaneously), which means that the memory required for learning scales linearly with the number of neurons.
Prior work has demonstrated the effectiveness of three-factor learning with SNNs on neuromorphic hardware, with the third factor drastically improving learning by projecting task-specific errors to the neurons \cite{Baldi_etal17_learmach,Stewart_etal20_onchfews, Stewart_etal20_onlifews}.

(iii) The exact analytical form of the weight gradients assumes that weight updates are performed continuously in time. In a typical digital implementation, this translates into writing memories at every time step, which is both energy-inefficient and slow. 
To solve this problem, \ac{SOEL} operates in an \textit{error-triggered} manner: weight updates occur only when an accumulated error threshold is crossed \cite{Payvand_etal20_errothre}. 
This drastically reduces the number of updates necessary for learning, while largely preserving the learning performance. 
More specifically, for a given input and target pair, the error $e_i$ for neuron $i$ is computed with the post-synaptic neuron using the following equation:
\begin{equation}
\label{eq:error}
\begin{split}
    & e_i(t) = s^*_i - \bar{s}_i(t), \\
    & y_i(t) = \begin{cases} 
                0,\text{ if $-\theta > e_i(t) > \theta$ }\\
                e_i(t), \text{otherwise}\\
                \end{cases}
\end{split}
\end{equation}
where $s^*_i$ is the target number of spikes during the time window $T$ for this neuron, and $\bar{s}_i(t) = \sum_{t-T}^t s_i(t)$ is the number of spikes produced by neuron $i$ in the previous $T$ timesteps.
$T$ is the interval during which errors are computed to determine if the error reaches the threshold $\theta$ for plasticity to occur.
The computation of $y_i$  is carried out on a general purpose processor \cite{Payvand_etal20_oncherro}.
The Intel Loihi includes general purpose x86 ``embedded cores'' for such purposes. 
The embedded processor calculates the error term $e_i(t)$ and the thresholded error term $y_i(t)$, and writes the latter to a post-synaptic state on the Loihi chip's plasticity state to update the synaptic weights.
The embedded processor induces a higher latency than the plasticity dynamics. 
To mitigate this, we compute and apply $\Delta w_{ij}$ every $T$ steps. 
Furthermore, if the error is below the threshold $\theta$, then no update is applied.
Because post-synaptic states on the chip can only be positive, in practice the $e_i(t)$ is offset by the constant term which is then canceled out in the plasticity dynamics (see Methods).
The following equation describes \ac{SOEL}:
\begin{equation}\label{eq:soel}
\Delta w_{ij} = \eta p_j(T)y_i(T).
\end{equation}
Here, $\eta$ is the learning rate, $p_j(t)$ is a pre-synaptic trace variables that reflects a non-weighted presynaptic potential, as required by gradient-based learning in \acp{SNN} \cite{Neftci_etal19_surrgrad}. 
Since we use current-based \ac{LIF} synapses, \ac{SOEL} with the pre-synaptic traces $p_j(t)$ are second-order linear filters.
The postsynaptic term $\sigma'$ present in \refeq{eq:loss} is omitted from \refeq{eq:soel}, \emph{i.e.} it is always equal to 1. 
This is for two reasons: firstly the first generation of the Loihi chip (Loihi 1) did not allow three factors. Secondly, \ac{SOEL} is used for learning in the final layer and omitting this term does not affect the sign of the gradients in this layer.
\ac{SOEL} is an approximation of the analytically computed weight gradients, and thus may fail to reduce the error to zero.
We emphasize here that although the loss function is rate based, the learning is inherently spike-based. This is because the individual spikes contribute to the neural and synaptic traces, enabling the neuron to learn patterns of spike in a way similar to surrogate gradient learning \cite{Neftci_etal19_surrgrad}.
We further note that the use of rate-based losses with surrogate gradients is the common practice in the field \cite{Shrestha_Orchard18_slayspik,Kaiser_etal19_synaplas,Zenke_Neftci21_brailear}.
\subsection{Bi-level Optimization algorithm and Meta Learning Algorithms}
We describe here the bi-level optimization procedure.
We meta-train models using surrogate gradient MAML similar to MAML SNN \cite{stewart2022meta}.
The MAML based framework is used here to learn a parameter initialization $w_0$, such that from the initialization one-shot adaptation to new samples can occur in the neuromorphic hardware.
To accomplish this, datasets are split into three sets of meta-tasks:  meta training $\mathcal{T}^{trn} $, meta validation $\mathcal{T}^{val} $, and meta testing $\mathcal{T}^{tst} $. 
Each task $T$ of meta-task $\mathcal{T}$ consists of a training dataset $\mathcal{D}^{trn}$ and a test dataset $\mathcal{D}^{tst}$ of the form $\mathcal{D}=\{x_i,t_i\}_{i=1}^M$.
Here $x_i$ denotes the input data, $t_i$ the target (label) and $M$ is the number of target samples.

Similar to MAML, the parameter initialization is optimized with bi-level optimization using two nested optimization loops, one ``inner'' loop and one ``outer'' loop. 
The inner loop imitates the online \ac{SOEL} dynamics, using the factors derived from the \ac{CUBA} neuron (\refeq{eq:cuba_lif}) and the Cross Entropy loss.
\begin{equation}\label{eq:inner_loop}
\begin{split}
    w_{n+1}(w_n, \mathcal{D}^{trn}) = w_{n} - \alpha \Delta w (s, w_n), \\
    \text{where }\{x,t\}\in\mathcal{D}^{trn}\text{ for }n=1,...,N.
\end{split}
\end{equation} 
Here $\Delta w (x, w_n)$ is the \ac{SOEL} update with $N$, the number of adaptation steps of the inner loop, and $\alpha$ is the inner loop learning rate. 
The dependence of $\Delta w (x, w_n)$ is made explicit for the outer loop gradients below.
Furthermore, the inner loop operations are traced for auto-differentiation \cite{Griewank_Walther08_evalderi}, such that their gradients can be seamlessly computed in the outer loop. 
During learning, a task is sampled from the training meta-task $\mathcal{T}^{trn}$ and $N$ inner loop updates are made using batches of data sampled from $\mathcal{D}^{trn}$. 
The resulting parameters $w_{N}$ are then used to update the outer loop using the matching test dataset $\mathcal{D}^{tst}$.

The parameters updated by the gradients calculated in the inner loop are then input to the outer loop where they are evaluated on the validation set for the second gradient calculation, again using cross entropy loss but instead of being optimized with stochastic gradient descent the outer loop is optimized with ADAM \cite{kingma2017adam}. 
The loss function optimized in the outer loop loss is:
\begin{equation}
    \mathcal{L}_{outer}(w_0) = \sum_{T \in P(\mathcal{T}^{trn})} \mathcal{L}(f(w_N(w_0, \mathcal{D}^{trn}), \mathcal{D}^{tst})),
\end{equation}
where $T=\{\mathcal{D}^{trn}, \mathcal{D}^{tst}\}$. 
In practice, the above expression is generally computed over a random subset of tasks rather than the full set $\mathcal{T}^{trn}$.
Notice that the outer loop loss is computed over the test dataset $\mathcal{D}^{tst}$, while $w_N(w_0)$ is computed using the training dataset $\mathcal{D}^{trn}$, which is shown to improve generalization.

An ADAM optimizer is used in the outer loop training \cite{kingma2017adam}. 
Although ADAM incurs larger memory and compute resources compared to \ac{SGD} it does not impact the deployment on the hardware which is solely based on \ac{SOEL}. 

Since digital neuromorphic hardware typically uses integer weights, full precision ``shadow weights'' \cite{Shrestha_Orchard18_slayspik} are quantized and used during the forward inference phase of the inner and outer loops.
During the backward pass, full precision gradients and weights are used to update parameters.
\subsection{Experimental Setup and Datasets}\label{sec:Datasets}
Few-shot learning aims to achieve rapid adaptation when prior knowledge related to the task domain is available.
Rather than iterating many times over many samples of data to learn, few-shot learning utilizes prior knowledge of the task domain to learn from one or a few samples of the data.
To accomplish this, we set up N-way K-shot learning experiments.
Here, $N=5$ is the number of classes to be classified and $K=1$ is the number of shots, or sample presentations, given to the model for learning.

\begin{figure}[h!]
    \includegraphics[width=.7\linewidth]{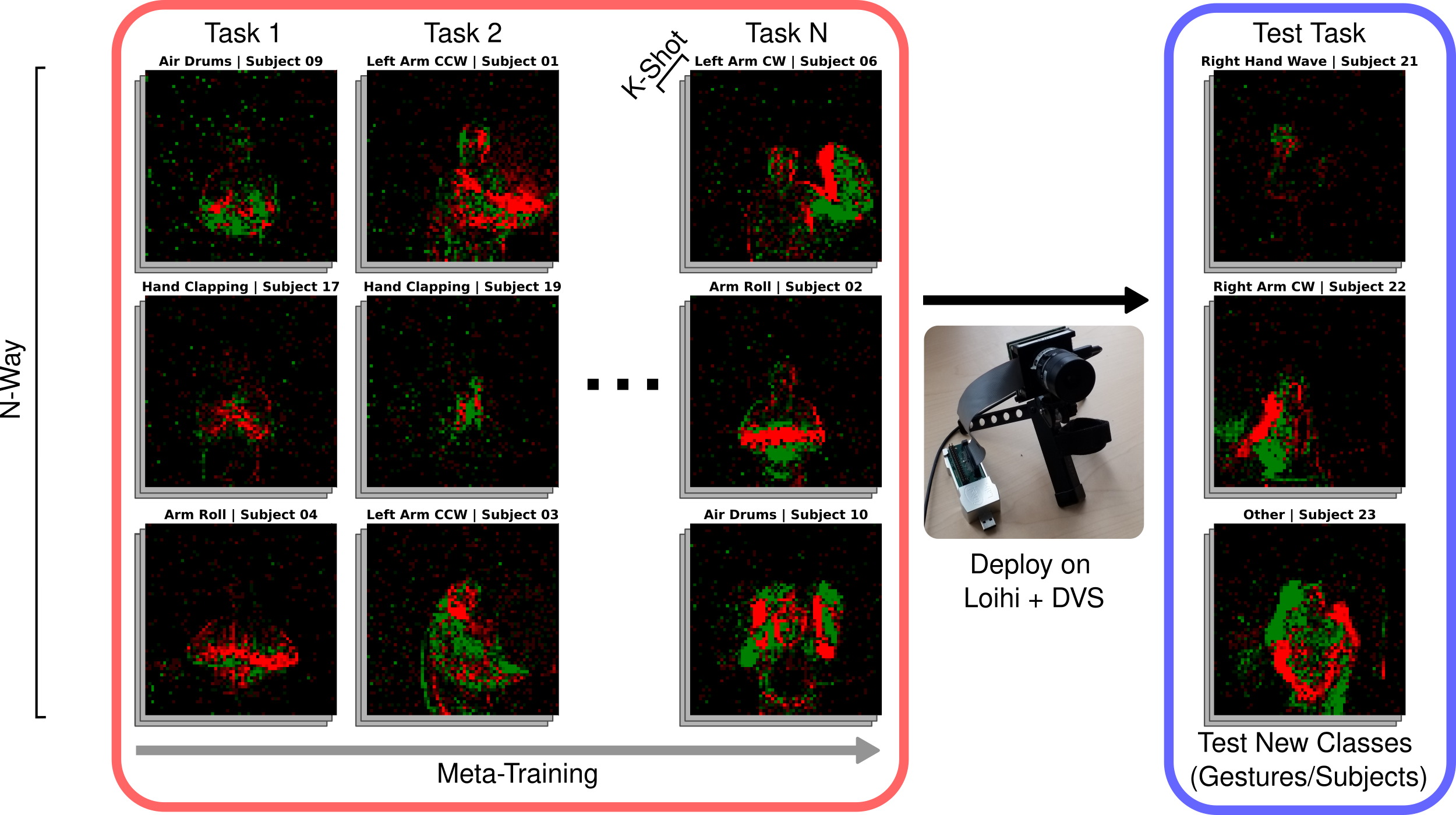}
    \centering
    \caption{Examples of each of N-shot K-way meta training on the DVS Gestures dataset. The network is trained on K-shots of various classes (gestures) of the training dataset, meaning that $K$ recordings samples are used to learn an N-way classification task. This is repeated until convergence. Then, the meta-trained model is tested by training on $K$ samples of each unseen class and subject. For visualization purposes here, the images are DVS events averaged over $300$ms into frames, however the learning neural network operated at a time step of $1$ms for 100$ms$ samples.}
    \label{fig:double_nmnist} 
\end{figure}
%

We evaluated our models on modified datasets captured using event-based vision sensors, including the neuromorphic MNIST (N-MNIST), the American Sign Language Dynamic Vision Sensor (ASL-DVS) and the DvsGesture datasets\cite{Lichtsteiner_etal08_128x120,Posch_etal11_qvga143,Orchard_etal15_convstat,bi2019graph,Amir_etal17_lowpowe}.
Each meta dataset consists of 32x32x100 ms event data sample and are constructed as follows:
\begin{itemize}
    \item Double N-MNIST: From the 10 class N-MNIST datasets' 60k training and 10k test samples, Double N-MNIST horizontally combines two N-MNIST samples, resulting in 100 classes. These 100 classes were divided into 64 meta-train, 16 meta-validation, and 20 meta-test classes.
    \item Double ASL-DVS: Similarly to Double N-MNIST, the Double ASL-DVS' concatenates the 24 classes (A-Y excluding J), 4200 samples per class, resulting in 576 tasks. There were split in 369 meta-train, 92 meta-validation, and 115 meta-test classes.
    \item DvsGesture subject+class: 29 individuals, 10 gestures, 1078 training, and 264 test samples. We created splits that used both gesture and individual, resulting in 319 tasks. These were split in 253 meta-train, and 66 meta-test classes \cite{Lichtsteiner_etal06_1281120d,yik2023neurobench}. Examples of images from the DvsGesture dataset and how it is implemented for meta-training and meta-testing are shown in \reffig{fig:double_nmnist}. The task here is to classify the combination of subject and gesture category.
\end{itemize}

To meta-train models compatible with the Intel Loihi, we trained SNNs with Intel's Lava-DL open source library \cite{orchard2021efficient}, which is the first stage of the meta-learning procedure.
Lava-DL is a deep library to facilitate offline training and online training and inference within the Lava framework.
The library includes a functional differentiable simulator of the Intel Loihi's neuron dynamics, and quantization-aware training with the same stochastic rounding and precision used in the Loihi chip.
Therefore, models meta-trained with Lava-DL are transferable for use with Intel's Loihi hardware.

\begin{figure}
    \centering
    \includegraphics[width=1.0\linewidth]{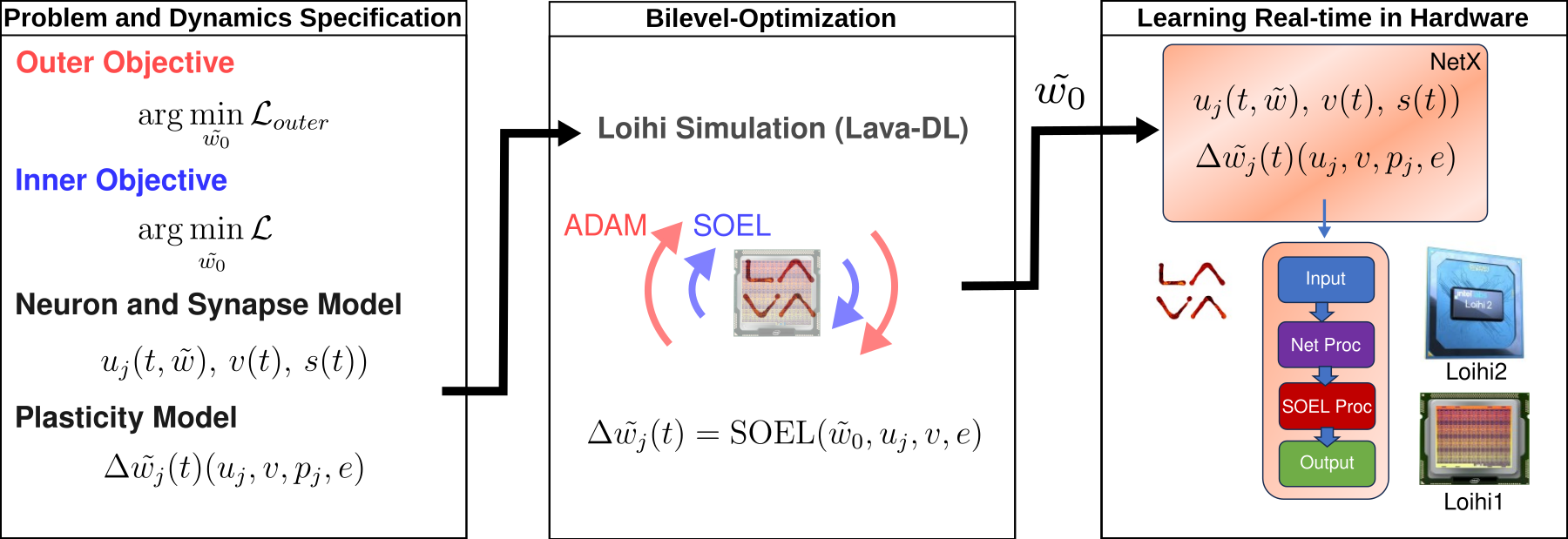}
    \caption{{Pipeline for training the Intel Loihi on few-shot learning}.
             (Left) The outer and inner objectives are first defined as differentiable functions of the neuron, synapse, and plasticity models. (Middle) These objectives are optimized using a Loihi simulation based on the Lava-DL library, which includes a functional simulator of the Loihi chip. Since Loihi uses integer precision weights, quantization-aware training is used where full precision shadow weights are quantized during the forward inference phase. Two gradient learning phases are performed, first in the inner loop (blue) and then in the outer loop with multiple shots of validation or test data (red). (Right) The resulting quantized parameters are transferred to the Intel Loihi neuromorphic hardware using Lava-DL NetX and Lava processes, enabling running on both the Loihi 1 and Loihi 2 in real time.}
    \label{fig:pipeline}
\end{figure}

The pipeline for meta-training models compatible with Intel's Loihi hardware for meta-testing on hardware is shown in \ref{fig:pipeline}.
For all results, we meta-trained models to perform 5-way 1-shot learning with Lava-DL.
A Lava-DL network is optimized across tasks using the ADAM optimizer \cite{kingma2017adam} in the outer loop and optimized to one-shot classify 5 tasks at a time using \ac{SOEL}.
The models can then be transferred to the Intel Loihi hardware for adaptation with \ac{SOEL} for online learning at the edge using NetX.
All SNN-based models used a two layer network of linear layers with 512 neurons each, plus an output layer consisting of 5 neurons.

\section{Results}
\subsection{SOEL rule in a Single Neuron}
\begin{figure}
    \centering
    \includegraphics[width=\textwidth]{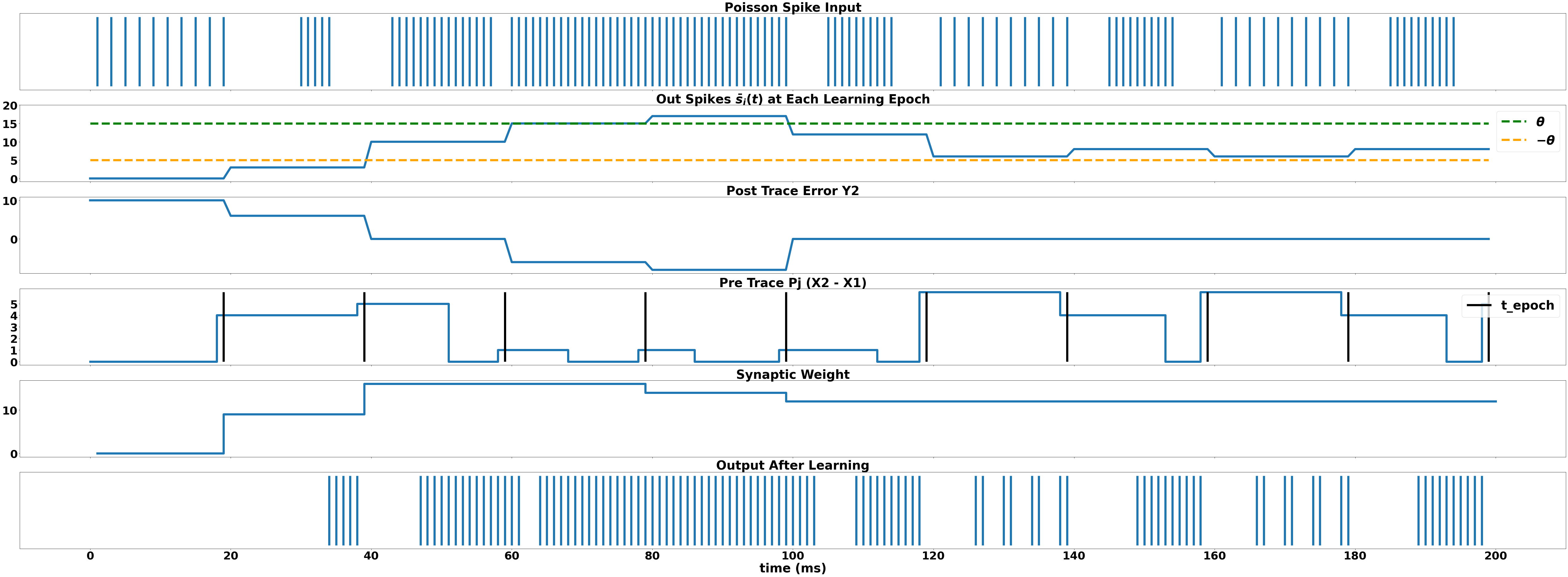}
    \caption{A single neuron example of the \ac{SOEL} learning rule. Given a Poisson spike input, \ac{SOEL} learns to regulate the output of the neuron to a target firing rate within a time interval when the error of the neuron is 0. If the error is not 0 then learning is triggered every $t_{epoch}$ and plasticity adjusts the synaptic weight to correct the firing rate of the neuron to the desired rate. The green and orange dotted lines on the out spike plot indicate the threshold $\theta$} 
    \label{fig:soel}
\end{figure}
We first show that the \ac{SOEL} rule operates as expected in a simple task aimed at adjusting the firing rate of a neuron towards a target value.
We show that in a single layer of \ac{LIF} neurons implemented on the Intel Loihi chip, \ac{SOEL} reduced the error to zero.
\reffig{fig:soel} demonstrates how \ac{SOEL} drives learning and reduces the error of a single neuron in the layer to zero, thereby achieving a target firing rate.
The single neuron example uses Intel's Lava\footnote{Lava is an open source neuromorphic programming framework available publicly at \url{https://github.com/lava-nc/lava}} software simulation of the Loihi 2 plasticity state machine.
In \reffig{fig:soel} the neuron stimulated by Poisson-distributed spike trains. 
These inputs elicit firing at a rate that is at first too high or too low.
Over time, \ac{SOEL} plasticity  adjusts the synaptic weight according to the error and trace values until it reaches the target firing rate.
In Intel Loihi chips, the learning step of the plastic synapse is executed at a set predetermined interval called $t_{epoch}$.
In the example shown in \reffig{fig:soel} $t_{epoch}$ is set to 20 and the occurrence of $t_{epoch}$ when the trace values are used for learning is marked with vertical lines.
The error is calculated using \refeq{eq:error} and if the error is not $0$ then learning is triggered and plasticity adjusts the synaptic weight to correct the neuron firing rate to the desired rate with \refeq{eq:soel}.


\subsection{Bi-level Optimization and Few-shot Learning in the Intel Loihi}
The bi-level optimization procedure aims to optimize the plastic network parameters hyperparameters to pre-train the network for fast online learning. 
In this bi-level optimization, \ac{SOEL} is effectively the inner loop adaptation, while the outer loop consists of a conventional gradient backpropagation-through-time across the network with \ac{SOEL}.

Here \ac{SOEL} is only enabled in the last layer of the network for the following reasons:
First, prior work demonstrated that MAML learns a suitable representation for few-shot learning instead of solely rapid learning in both ANNs \cite{Raghu2020Rapid} and SNNs \cite{stewart2022meta}, meaning networks with meta-learned initializations can achieve high performance on learning new tasks using only a single gradient update in the last layer.
The likely reason for this is that \ac{MAML} learns an adequate feature representation in the layers prior to the plastic ones \cite{Chen_etal19_modumeta}. 
Second, there is no requirement to spatially backpropogate errors in the inner loop, allowing for learning rules such as \ac{SOEL} to be implemented very efficiently in hardware.
Third, the outer loop optimization mitigates the approximations of \ac{SOEL} by optimizing the hyperparameters accordingly. 

We emphasize here that although \ac{SOEL} is used in the final layer, the bi-level optimization modifies the parameters of the entire network. 
The difference between the final layer synapses and the other layers is that the former undergo changes during the presentation of a new sample and thus specialize in the new task. 
In contrast, the synaptic weights in the other layers represent features that are common to all tasks in the meta dataset.
Although in some cases, optimizing the penultimate layer was reported to also benefit accuracy \cite{Chen_etal19_modumeta}, the results of the previous simulation showed that the expansion of the plasticity deeper into the network produced only marginal improvements \cite{Stewart_Neftci22_metaspik}.

We demonstrate one-shot learning on the Loihi neuromorphic hardware using \ac{SOEL} with a meta-trained network on the datasets described in the following sections.
Once the model is meta-trained (the ``first stage'') the network (the ``second stage'') is transferred to the Intel Loihi for online one-shot learning on the device.

\subsubsection*{Meta-Training for Few-Shot Learning}
To build the prior knowledge for effective few-shot learning in neuromorphic hardware, we optimize for few-shot learning on event-vision classification tasks.
We optimize the model for three different event-based vision tasks.
The Double NMNIST task \cite{stewart2022meta}, a double digit version of the Neuromorphic MNIST dataset \cite{Orchard_etal15_convstat}; The Double ASL-DVS \cite{stewart2022meta}, two American Sign Language letter gestures presented side-by-side recorded with the DAVIS 346 \cite{bi2019graph} event-vision sensor; and the DVSGesture dataset \cite{Amir_etal17_lowpowe}, a mid-air gesture dataset recorded with the DAVIS 128 \cite{Lichtsteiner_etal08_128x120} event-vision sensor. 
In our experiments with the DVSGesture dataset the task is to not only classify the gesture but the individual performing the gesture as well; therefore, there are 44 classes in total.
The details of how the datasets are prepared for the experiments and split for bi-level optimization can be found in Section \ref{sec:Datasets}.

We performed multiple few-shot bi-level learning experiments across the datasets.
The meta-test set accuracy results of our few-shot bi-level learning experiments are shown in Table \ref{tab:results}.
To test the models, each model is tested in a one-shot learning trial.
In a trial the model is first given a single shot of data to learn from in an inner loop gradient step, then from the single shot learned parameters inference is performed on 10-shots of the test data.
The accuracy values shown in Table \ref{tab:results} are averaged over 200 random trials using the same random seed
plus or minus the standard deviation across those trials.
Neuromorphic hardware has constraints that limit the network compared to a full precision \ac{SNN} trained in conventional GPU hardware.
These constraints are as follows: i) Quantization: the neuron and synapse states are quantized to fixed-point integer values with signed 8 bits of precision for the weights that are stochastically rounded to even values; ii) Hard reset: the reset of the neuron is hard meaning a neuron that spikes membrane potential will reset to 0; and iii) Plasticity processor: use of the SOEL plasticity rule.
The SNN model uses floating point precision by default with a soft reset (\emph{i.e.} the membrane potential is subtracted by a fixed amount after spiking), as opposed to the hard reset to a reset potential (used in the Intel Loihi).
Therefore we demonstrate how the performance is impacted by the difference in the neuron model's reset and quantization in Table \ref{tab:results}.
This progression aims to disentangle the impacts of the hardware realities.
While there is a general accuracy decrease shown by the SNN model as constraints are added, the performance is comparable to the Lava-DL trained models that are deployable on Intel's hardware.
The Lava-DL SOEL trained models were then transferred to Intel's Loihi 1 hardware where the last layer of the network were learned using SOEL.
The accuracies of the Loihi 1 models are comparable to the Lava-DL SOEL models shown in table \ref{tab:results}.
Additionally, the Lava-DL SOEL trained models were used to learn in the last layer of the models in Intel's Lava Loihi 2 hardware simulator to demonstrate the accuracy of models trained on Intel's latest neuromorphic hardware again with results comparable to the Lava-DL SOEL models. 

As a baseline, we compare the SNN meta-trained models to a k-nearest-neighbors (KNN) model.
KNNs use supervised learning to make predictions based on the similarity of a new data point to its k nearest neighbors in a training dataset.
We compare to a KNN model to demonstrate the importance of having a pre-training stage to build a knowledge base for accurate few-shot learning.
KNN models do not require a training phase and have been shown in prior work to have been effectively demonstrated for classification on Intel Loihi hardware \cite{frady2020knn}.
The KNN performs much worse on the 5-way 1-shot learning tasks than the meta-trained SNN models, particularly on the Double ASL and DvsGesture datasets with the KNN performance close to random on the DvsGesture dataset.
This suggests that training, or optimization, particularly bi-level optimization is necessary to solve few-shot learning.

Table \ref{tab:loihi} shows energy, time, and power statistics for learning one sample of data using \ac{SOEL} from the meta-trained model. 
For each sample, meta-trained models on Loihi using \ac{SOEL} for one-shot learning use less than 1mW of dynamic power and take only about 10-20ms to learn a sample making one-shot learning with Loihi using our methods are both accurate and power efficient.

\begin{table}[]
\caption{One-shot Learning Accuracy Comparison}
\label{tab:results}
\centering
\begin{tabular}{l|l|l|l}
\hline
Dataset                                 & Platform                                  & Method                                                   & Test Accuracyx  \\ \hline
\multirow{9}{*}{Double NMNIST}          & \multirow{4}{*}{GPU}                      & MAML SNN \cite{stewart2022meta}                          & $98.2\pm 1.1$\% \\ 
                                        &                                           & MAML SNN+hard reset                                      & $94.7\pm1.6$\%  \\ 
                                        &                                           & MAML SNN+quantized                                       & $97.7\pm1.0$\%  \\ 
                                        &                                           & MAML SNN+hard reset+quantized                            & $91.5\pm2.4$\%  \\ \cline{2-4} 
                                        & \multirow{2}{*}{Loihi 1 Sim}              & MAML SNN+hard reset+quantized                           & $91.1\pm2.5$\%  \\ 
                                        &                                           & MAML SNN SOEL                                        & $90.5\pm2.1$\%  \\             
                                        & Loihi 1 HW                                & MAML SNN SOEL                      & $93.3\pm1.3$\%  \\             
                                        & Loihi 2 Sim                               & MAML SNN SOEL              & $95.6\pm1.3$\%  \\ \cline{2-4} 
                                        & CPU                                       & KNN                                                      & $83.0\pm9.8$\%  \\ \hline
\multirow{9}{*}{Double ASL}             & \multirow{4}{*}{GPU} & MAML SNN \cite{stewart2022meta}                          & $96.0\pm2.3$\%  \\             
                                        &                      & MAML SNN+hard reset                                      & $93.2\pm2.2$\%  \\             
                                        &                      & MAML SNN+quantized                                       & $91.4\pm2.1$\%  \\             
                                        &                      & MAML SNN+hard reset+quantized                            & $92.4\pm2.0$\%  \\ \cline{2-4} 
                                        & \multirow{2}{*}{Loihi 1 Sim}              & MAML SNN+hard reset+quantized                            & $92.1\pm2.6$\%  \\  
                                        &                                           & MAML SNN SOEL                                            & $91.8\pm2.8$\%  \\  
                                        & Loihi 1 HW                                & MAML SNN SOEL                  & $91.3\pm3.6$\%  \\  
                                        & Loihi 2 Sim                               & MAML SNN SOEL              & $94.7\pm1.4$\%  \\ \cline{2-4} 
                                        & CPU                                       & KNN                                                      & $39.6\pm9.6\%$  \\ \hline
\multirow{9}{*}{\makecell{DvsGesture\\Subject+Class}} & \multirow{4}{*}{GPU}          & MAML SNN                                     & $95.1\pm1.6$\%  \\ \cline{3-4} 
                                        &                                           & MAML SNN+hard reset                                      & $96.9\pm1.3\%$  \\  
                                        &                                           & MAML SNN+quantized                                       & $95.6\pm1.4$\%  \\  
                                        &                                           & MAML SNN+hard reset+quantized                            & $95.1\pm2.0$\%  \\ \cline{2-4}  
                                        & \multirow{2}{*}{Loihi 1 Sim}              & MAML SNN+hard reset+quantized                            &  $95.7\pm1.0$\%\\ 
                                        &                                           & MAML SNN SOEL                                        & $95.8\pm0.8$\% \\ 
                                        & Loihi 1 HW                                & MAML SNN SOEL                      & $97.1\pm2.0$\% \\ 
                                        & Loihi 2 Sim                               & MAML SNN SOEL                  & $96.7\pm1.0$\% \\ \cline{2-4} 
                                        & CPU                                       & KNN                                                      & $23.2\pm4.7\%$ \\ \hline
\end{tabular}
\end{table}

\begin{table}[t]
\caption{Loihi 1 One Sample Learning Stats}
\label{tab:loihi}
\centering
\begin{tabular}{llllll}
Dataset                             & Energy (mJ)            & Time (ms)             & Dynamic Power (mW) & Total Power (mW)  & Cores                 \\ 
               
\hline
Double NMNIST                       & 9.34$\pm$1.84                  & 10.57$\pm$2.09                 & 0.51$\pm$0.07        & 15.97$\pm0.09$               & 14 \\ 
Double ASL                          & 22.67$\pm$3.45                  & 25.67$\pm3.79$                  & 0.55$\pm$0.04   & 15.72$\pm0.04$                 & 14 \\ 
DvsGesture Subject                    & 13.71$\pm$1.91                  & 15.44$\pm$2.26                 & 0.75$\pm$0.08  &          16.10$\pm0.11$            & 26 \\ 
\multicolumn{6}{p{0.95\linewidth}}{\tiny $^*$Intel Loihi measurements obtained using NxSDK v1.0.0 running on on a single chip Loihi 1 Kapoho-Bay system (USB form factor). Performance results are based on testing as of Jan 2024 and may not reflect all publicly available security updates. Results may vary. 
} \\
\end{tabular}
\end{table}

\pagebreak
\section{Discussion}
Neuromorphic hardware equipped with synaptic plasticity utilizes event-driven weight updates to rapidly adjust to new tasks.
Due to the mostly local and sparse nature of neuromorphic hardware, it is particularly well suited for online learning at the edge.
We adopted a powerful meta-learning strategy that can adapt three-factor plasticity rules for accurate online one-shot adaptation at the edge.

Our work differs from previous efforts to learn from scratch with neuromorphic hardware \cite{Detorakis_etal18_neursyna,Frenkel_Indiveri22_reck28nm,Cramer_etal22_surrgrad,Chicca_etal13_neurelec}. 
Although such approaches work in theory and well controlled learning procedures, they did not provide a viable learning technology.
This is because online learning from streaming data in real-world scenarios has several challenges which are not addressed in prior work: data is not i.i.d., training is lengthy, and impractically small learning rates are necessary to average across the entire dataset.

We address these challenges using a combination of offline and online learning.
Offline pre-training takes place on high-power, high-throughput hardware with access to extensive datasets allows for the development of initial models that capture common features in the task domain.
Gradient-based meta-learning such as MAML automates this pre-training.
These pre-trained models can then be fine-tuned or adapted on the device using online learning techniques, such as the \ac{SOEL}.
This hybrid approach strikes a balance between leveraging prior knowledge and adapting to real-time data, offering improved robustness and faster convergence.

Furthermore, when meta-training a network with plasticity dynamics, the resulting pre-trained model can mitigate the non-idealities of the adaptation rule with respect to gradient descent, at least for the selected number of training steps.
This observation means that certain challenging hardware constraints involved in the implementation of the synaptic plasticity rule can be relaxed without significantly losing performance. 
In the case of \ac{SOEL}, we relaxed the update frequency and the nature of the post-synaptic factor. 
Other related work mitigated the downsides of \ac{STDP} \cite{miconi2018plasticnets} and the non-idealities in hardware \cite{Yu_etal22_traiwith}.

Using bi-level optimization to enable fast learning with brain-like synaptic plasticity has previously been attempted \cite{miconi2018plasticnets,rosenfeld2021fast}.
Additionally, other work demonstrated the benefits of bi-level optimization for improving the data and power efficiency of learning at the edge \cite{rosenfeld2021fast}.
For example, \cite{miconi2018plasticnets} showed that plasticity can be optimized by gradient descent, or meta-learned, in large artificial networks with Hebbian plastic connections, called differentiable plasticity.
Our work is closest to \cite{wu2020braininspired}: Building on the differentiable plasticity, they meta-optimized Hebbian-like \ac{STDP} learning rules and local meta-parameters on the event-based datasets for few-shot learning tasks using a model of the Tianjic hardware.
Synapses store two components, one that is optimized during bi-level optimization and one that is updated using \ac{STDP} like plasticity.
This is similar in spirit to optimizing initial weights as in \ac{MAML} and in our work.
Our approach differs in three ways: (i) by using a three-factor synaptic plasticity rule (\emph{i.e} the \ac{SOEL}), which takes error signals computed from a custom loss function thus enabling the learning of supervised and self-supervised approaches, (ii) we demonstrate on-line learning in custom digital neuromorphic learning hardware, and (iii) by operating the neural dynamics on finer time scales ($1\mathrm{ms}$ time steps compared to $12.5\mathrm{ms}$ time steps in the case of \cite{wu2020braininspired}, which is closer to conventional frame camera rates), providing more than an order of magnitude in temporal resolution.

We demonstrated \ac{SOEL} and the associated bi-level optimization in a range of event-based classification tasks, with data recorded from event-based vision sensors.
The class of few-shot learning problems with event-based data has been identified as a key benchmark and application by the broader neuromorphic engineering community \cite{yik2023neurobench}.
While our work focused on supervised learning with visual data, our method and general approach is applicable to other supervised learning tasks.
In fact, the MAML framework has been established for tasks such as reinforcement learning \cite{beck2023metarlsurvey}, multilingual speech emotion classification\cite{naman2022mamlspeech}, and self-supervised learning \cite{Lin2021selfmeta}.
In this work, \ac{SOEL} was limited to the final layer of the network. 
This is not a conceptual limitation of \ac{SOEL} itself, since similar error-triggered learning using local losses was already demonstrated in prior work \cite{Payvand_etal20_oncherro}. 
Rather, this choice was made because MAML is known to learn a suitable representation for few-shot learning instead of rapid learning \cite{Raghu_etal19_rapilear}.
Indeed, simulation results show that extending the plasticity deeper in the network yields marginal improvements \cite{Stewart_Neftci22_metaspik}.

One limitation of \ac{SOEL} is for continual learning: Catastrophic forgetting of previous knowledge will occur after multiple online learning epochs. 
Looking forward, we envision two approaches to solve this problem: (i) Federated learning and (ii) Continual learning techniques, which we detail below.
First, federated learning is dedicated to addressing the problem of sharing knowledge from edge devices to a cloud model, while preserving privacy, communication and accuracy \cite{Li_etal20_fedelear,McMahan_etal17_commlear,Venkatesha_etal21_fedelear,Yoon_etal21_fedecont}.
Using meta-learning approaches, the meta-learned models can also be fine-tuned with the data collected at the edge \cite{Fallah_etal20_persfede}.
Online meta-learning \cite{Finn_etal19_onlimeta,Rosenfeld_etal21_fastonde,Javed_White19_metarepr,Denevi_etal19_onlimeta} can enable continual meta-learning in the offline loop, provided task-relevant information can be transferred back to the offline agent.
Secondly, currently developed methods to use continual learning can be implemented \cite{Kudithipudi_etal23_desiprin}. These consist of dynamic architectures, whereby branches of the network are modulated for example using neuromodulation, metaplasticity where synapses are equipped with internal consolidation variables and replay methods.

Another limitation is general to the meta-learning problem: the pre-trained models generalize poorly to tasks that are not present in the meta-training set. 
An example of such a new task is flipping the event-camera inputs upside down. In such cases, performance drops drastically.
Using larger datasets and models is a natural solution to this problem. 
This is the case with so-called foundational models \cite{Bommasani_etal21_opporisk} that demonstrate downstream task performance that sometimes even exceeds those of models specifically trained on such tasks \cite{Radford_etal21_leartran}. 
Although the scale of current digital neuromorphic hardware and datasets are far from those used in foundational models, this gap is closing thanks to a better selection of the dataset and a simplification of foundational models \cite{zhu2023spikegpt} and the development of large-scale neuromorphic hardware.

Bi-level optimization on can be used to achieve high performing rapid learning on ultra-low power neuromorphic hardware.
While our method with MAML and SOEL does not solve the problem of catastrophic forgetting, future work using federated and continual learning techniques could help prevent catastrophic forgetting.
Additionally, ongoing work on foundation models and the development of large-scale neuromorphic hardware could benefit from online rapid adaption through bi-level optimization across a multitude of task domains such as autonomous vehicles, and speech translation. 
We expect that such meta-learning approaches will lead to a substantial redefinition of synaptic plasticity requirements for neuromorphic processors and pave the way for exciting and novel learning applications at the edge.

\section*{Acknowledgements}
This research was supported by the Intel Corporation (KS, SS), by the Federal Ministry of Education, Germany (project NEUROTEC-II grant no. 16ME0398K and 16ME0399), and by Neurosys as part of the initiative "Cluster4Future" funded by the Federal Ministery of Education and Research BMBF (03ZU1106CB) (EN).

\section*{Data Availability}
The data that supported the findings of this study are found at 
https://github.com/nmi-lab/torchneuromorphic.

\section*{Code Availability}
The software source code to reproduce the primary results is available at 
https://github.com/nmi-lab/snn\_maml and https://github.com/kennetms/lava-dl.

\section{Appendix}
\subsection{Implementation of \ac{SOEL} on the Intel Loihi 1 and 2 Plasticity State Machine}
\ac{SOEL} implemented in the Intel Loihi chip follows Eq.~(\ref{eq:soel}), with the difference that a constant factor is introduced to represent negative errors. 
This is because a post-synaptic trace variable is used as a placeholder for $e_i(t)$. Because post-synaptic variables cannot be negative, we offset the values with a constant $c$ as follows:
\begin{equation}
\label{eq:error_alg}
\begin{split}
    e_i(t) & = s^*_i - \bar{s}_i(t), \\
    y_i(t) & = \begin{cases} 
                c + e_i(t),\text{ if $-\theta > e_i(t) > \theta$ }\\
                0, \text{otherwise}\\
                \end{cases}\\
    \Delta w_{ij} &= \eta p_j(T)(y_i(T)-c).
\end{split}
\end{equation}
Furthermore, the second order trace $p_j$ is computed using a difference of two first order traces (indices $j$) omitted:
\begin{equation}
    \begin{split}
        p(t)   &= x^2_j(t) - x^1(t) \\
        x^1(t) &= \alpha_u x^1(t-1) - (1-\alpha_u) s(t) \\
        x^2(t) &= \alpha_v x^2(t-1) - (1-\alpha_u) s(t) \\
    \end{split}
\end{equation}
with $\alpha_v>\alpha_u$.

On the Loihi 1 hardware, custom programs run on the Loihi's x86 embedded processors to calculate the error to be written to the post-synaptic variable of the learning rule.
The programs are run every learning epoch, with a predetermined timestep interval input to the learning rule whose value can be in the range \{0,...,63\}.
Therefore, multiple updates will typically be performed depending on the number of timesteps of a sample.
During a learning epoch the error is calculated as follows.
First is the determination of if the learning epoch occurs during a sample presentation or a blank time.
A blank time is a period of time in which no sample is presented which can allow traces and neuron parameters to decay back to 0 before presenting another sample.
If the learning epoch occurs during a blank time then 0 is written to the post trace and the error is not calculated.
Otherwise, a check is performed to determine if learning should occur or simply inference.
This is done by checking if a label for the class that should be updated is given in which case an error for that class should be calculated and written to the post-trace and otherwise no learning should be done.
If a label is given and the learning epoch does not occur during a blank time then the error will  be calculated following \refeq{eq:error} to allow the learning rule to update the synaptic weights.
From \refeq{eq:error_alg}, $s^*_i$ is the target spike rate between learning epochs, $\bar{s}_i(t)$ is the actual output spike count between learning epochs by the neuron that should learn to predict the class. 
The error is only written to the post-trace if it is greater than a threshold; otherwise 0 will be written to the post-trace and no learning will occur. 
The error is multiplied by the second-order trace $p_j$ and the learning rate $\eta$ to calculate the change in synaptic weight at each learning epoch.

In addition to the Loihi 1 hardware implementation, we tested \ac{SOEL} using a software simulation of the Loihi 2 plasticity using Intel's Lava open source software library.
The same plasticity dynamics used by Loihi 1 are implemented by Loihi 2, therefore \ac{SOEL} is implementable on Intel's newer Loihi 2.
The Loihi 2 implementation used to obtain the results shown in \reftab{tab:results} and \reffig{fig:soel} is algorithmically identical to the Loihi 1 hardware implementation.
The plasticity state machine of the Loihi 2 hardware has greater flexibility and more features than the Loihi 1, which will be explored in future work.

\subsection{Intel Loihi Neuron Model and Synaptic Plasticity}
The Intel Loihi is a neuromorphic processor equipped with a diverse array of features including dendritic compartments, hierarchical connectivity, synaptic delays, and programmable synaptic learning rules \cite{Davies_etal18_loihneur, orchard2021efficient}.
\acp{SNN} on Intel's Loihi 1 uses Current Based (CUBA) Leaky Integrate \& Fire (LIF) neuron dynamics and the synaptic weights are integers stochastically rounded to 8-bits of precision signed with a step of two allowing values in the range of $\{-256, -254, \cdots, 252, 254\}$.
Intel's Loihi 2 supports neuron models beyond the CUBA LIF model, but for consistency with the Loihi 1 we use the CUBA LIF model for our Loihi 2 experiments as well.
 The learning rule for a synaptic weight in the Intel Loihi 1 and 2 take the following sum-of-products form as follows \cite{Davies_etal18_loihneur}:
 \begin{equation}\label{eq:loihi_rule}
     \Delta w = \sum_k C_{k}\prod_l V_{kl},
 \end{equation}
 where  $w$ represents the synaptic weight variable assigned to the updated neuron pair from the source to the destination. $C_k$ is a scaling constant, and $V_{kl}[t]$ can be programmed to represent different state variables. These variables can be pre-synaptic spikes or traces, and post-synaptic spikes or traces. 
 The traces are represented using first-order linear filters and represent a form of synaptic eligibility similar to the traces in \ac{STDP} \cite{Bi_Poo98_synamodi,Sjostrom_etal08_dendexci}.

\subsection{Details to the Neuron Model}

Current-based neurons are commonly used in digital neuromorphic hardware thanks to its simplicity.
The \ac{CUBA} neuron dynamics are described as follows:
\begin{equation}\label{eq:cuba_lif}
\begin{split}
    u_j(t) &= \alpha_u u_j(t-1) + (1 - \alpha_u)\, w_j x_j(t) , \\
    v(t) &= \alpha_v v(t-1) + (1 - \alpha_v)\,\sum_j u_j(t), \\
    s(t) &= v(t) \geq \vartheta, \\
    v(t) &= v(t)\,(1-s(t)) 
\end{split}
\end{equation} 
where $u(t)$ is the synaptic current at time $t$, $\alpha_u$ is the decay constant for the current, $x(t)$ is the input at time $t$, $v(t)$ is the membrane potential at time $t$, $s(t)$ is a post-synaptic spike, and $\vartheta$ is the threshold at which the neuron fires a spike.
\ac{SOEL} is computed using the pre-synaptic trace $p_j(t)$, which is analytically computed from $\frac{\partial v(t)}{\partial w_j}$.
It is then trivial that $p_j(t)$ is equivalent to two first order filters applied to $x_j(t)$:
\begin{equation}\label{eq:cuba_lif2}
\begin{split}
    q_j(t) &= \alpha_u q_j(t-1) + (1 - \alpha_u)\, x_j(t) , \\
    p_j(t) &= \alpha_v p_j(t-1) + (1 - \alpha_v)\, q_j(t), \\
\end{split}
\end{equation} 
%
Training SNNs with gradient-based learning typically involves using Back Propagation Through Time (BPTT).
The dynamics in Eq.~\ref{eq:cuba_lif} cannot be used with gradient descent because the spike function $s(t) = v(t) \geq \vartheta$ is not differentiable.
The surrogate gradient approach approximates the spiking threshold function for gradient estimations with a smooth function. 


\bibliography{biblio_unique_alt, extra_bib, bib}
\bibliographystyle{IEEEtran}

\end{document}